\documentclass[10pt,twocolumn,letterpaper]{article}

\usepackage{cvpr}
\usepackage{times}
\usepackage{epsfig}
\usepackage{graphicx}
\usepackage{amsmath}
\usepackage{amssymb}
\usepackage{algorithm}
\usepackage{algorithmic}

\newcommand{\norm}[1]{\left\lVert#1\right\rVert}
\usepackage{multirow}
\usepackage{verbatim}
\usepackage{color}

\usepackage[pagebackref=true,breaklinks=true,letterpaper=true,colorlinks,bookmarks=false]{hyperref}

\cvprfinalcopy % *** Uncomment this line for the final submission

\def\cvprPaperID{****} % *** Enter the CVPR Paper ID here

\ifcvprfinal\fi
\begin{document}

%%%%%%%%% TITLE
\title{3DSSD: Point-based 3D Single Stage Object Detector}

\author{Zetong Yang$^{\dag}$\\
\and
Yanan Sun$^{\ddagger}$
\and
Shu Liu$^{\dag}$
\and 
Jiaya Jia$^{\dag}$
\\
\and
$^{\dag}$The Chinese University of Hong Kong~~~~~~$^{\ddagger}$The Hong Kong University of Science and Technology\\
\vspace{-2mm}
{\tt\small \{tomztyang, now.syn, liushuhust\}@gmail.com~~leojia@cse.cuhk.edu.hk} 
}

\maketitle
%\thispagestyle{empty}

%%%%%%%%% ABSTRACT
\begin{abstract}

  Currently, there have been many kinds of voxel-based 3D single stage detectors, while point-based single stage methods are still underexplored. In this paper, we first present a lightweight and effective point-based 3D single stage object detector, named 3DSSD, achieving a good balance between accuracy and efficiency. In this paradigm, all upsampling layers and refinement stage, which are indispensable in all existing point-based methods, are abandoned to reduce the large computation cost. We novelly propose a fusion sampling strategy in downsampling process to make detection on less representative points feasible. A delicate box prediction network including a candidate generation layer, an anchor-free regression head with a 3D center-ness assignment strategy is designed to meet with our demand of accuracy and speed. Our paradigm is an elegant single stage anchor-free framework, showing great superiority to other existing methods. We evaluate 3DSSD on widely used KITTI dataset and more challenging nuScenes dataset. Our method outperforms all state-of-the-art voxel-based single stage methods by a large margin, and has comparable performance to two stage point-based methods as well, with inference speed more than 25 FPS, 2$\times$ faster than former state-of-the-art point-based methods.
\end{abstract}

\section{Introduction}

In recent years, 3D scene understanding has attracted more and more attention in computer vision since it benefits many real life applications, like autonomous driving \cite{KITTIDATASET2} and augmented reality \cite{Multiple3Dtracking}. In this paper, we focus on one of the fundamental tasks in 3D scene recognition, \ie, 3D object detection, which predicts 3D bounding boxes and class labels for each instance within a point cloud.

Although great breakthroughs have been made in 2D detection, it is inappropriate to translate these 2D methods to 3D directly because of the unique characteristics of point clouds. Compared to 2D images, point clouds are sparse, unordered and locality sensitive, making it impossible to use convolution neural networks (CNNs) to parse them. Therefore, how to convert and utilize raw point cloud data has become the primary problem in the detection task.

Some existing methods convert point clouds from sparse formation to compact representations by projecting them to images \cite{MV3D,AVOD,MultiViewRandomForest,PedestrianDetectionCombine,Vote3Deep}, or subdividing them to equally distributed voxels \cite{VoxNet,VotetoVote,VOXELNET,YangLU18,yan2018second,lang2018pointpillars}. We call these methods voxel-based methods, which conduct voxelization on the whole point cloud. Features in each voxel are generated by either PointNet-like backbones \cite{POINTNET, POINTNET2} or hand-crafted features. Then many 2D detection paradigms can be applied on the compact voxel space without any extra efforts. Although these methods are straightforward and efficient, they suffer from information loss during voxelization and encounter performance bottleneck.

Another stream is point-based methods, like \cite{yang2018ipod,yang2019std,shi2018pointrcnn}. They take raw point clouds as input, and predict bounding boxes based on each point. Specifically, they are composed of two stages. In the first stage, they first utilize {\it set\ abstraction} (SA) layers for downsampling and extracting context features. Afterwards, {\it feature propagation} (FP) layers are applied for upsampling and broadcasting features to points which are discarded during downsampling. A 3D region proposal network (RPN) is then applied for generating proposals centered at each point. Based on these proposals, a refinement module is developed as the second stage to give final predictions. These methods achieve better performance, but their inference time is usually intolerable in many real-time systems. 

\paragraph{Our Contributions}

Different from all previous methods, we first develop a lightweight and efficient point-based 3D single stage object detection framework. We observe that in point-based methods, FP layers and the refinement stage consume half of the inference time, motivating us to remove these two modules. However, it is non-trivial to abandon FP layers. Since under the current sampling strategy in SA, \ie, furthest point sampling based on 3D Euclidean distance (D-FPS), foreground instances with few interior points may lose all points after sampling. Consequently, it is impossible for them to be detected, which leads to a huge performance drop. In STD \cite{yang2019std}, without upsampling, \ie, conducting detection on remaining downsampled points, its performance drops by about 9\%. That is the reason why FP layers must be adopted for points upsampling, although a large amount of extra computation is introduced. To deal with the dilemma, we first propose a novel sampling strategy based on feature distance, called F-FPS, which effectively preserves interior points of various instances. Our final sampling strategy is a fusion version of F-FPS and D-FPS.

To fully exploit the {\it representative points} retained after SA layers, we design a delicate box prediction network, which utilizes a candidate generation layer (CG), an anchor-free regression head and a 3D center-ness assignment strategy. In the CG layer, we first shift representative points from F-FPS to generate {\it candidate points}. This shifting operation is supervised by the relative locations between these representative points and the centers of their corresponding instances. Then, we treat these candidate points as centers, find their surrounding points from the whole set of representative points from both F-FPS and D-FPS, and extract their features through multi-layer perceptron (MLP) networks. These features are finally fed into an anchor-free regression head to predict 3D bounding boxes. We also design a 3D center-ness assignment strategy which assigns higher classification scores to candidate points closer to instance centers, so as to retrieve more precise localization predictions.

%and assign center-ness labels within range $[0, 1]$ to them,
%Another redundant design of point-based methods is multi-scale and multi-angle anchors because of large variance in different shapes and orientations of instances. In this paradigm, we treat each sampled point as the initial representation of instances, propose a center-aware layer to shift the sampled points and assign center-ness labels within range $[0, 1]$ to them, under the supervision of their relative locations to objects, which enables our model to realize the geometrical centers of objects. More details will be discussed in section \ref{ctr_aware_layer}. Finally, we use a modified SA layers to extract features of these translated points and generate their final predictions through an anchor-free regression head.

We eveluate our method on widely used KITTI \cite{KITTIDATASET1} dataset, and more challenging nuScenes \cite{nuscenes2019} dataset. Experiments show that our model outperforms all state-of-the-art voxel-based single stage methods by a large margin, achieving comparable performance to all two stage point-based methods at a much faster inference speed. In summary, our primary contribution is manifold.
% add time x 3

\begin{itemize}\vspace{-0.05in}

\item We first propose a lightweight and effective point-based 3D single stage object detector, named 3DSSD. In our paradigm, we remove FP layers and the refinement module, which are indispensible in all existing point-based methods, contributing to huge deduction on inference time of our framework.

\vspace{-0.05in}
\item A novel fusion sampling strategy in SA layers is developed to keep adequate interior points of different foreground instances, which preserves rich information for regression and classification.

\vspace{-0.05in}
\item We design a delicate box prediction network, making our framework both effective and efficient further. Experimental results on KITTI and nuScenes dataset show that our framework outperforms all single stage methods, and has comparable performance to state-of-the-art two stage methods with a much faster speed, which is 38ms per scene.
\end{itemize}

\section{Related Work}

\paragraph{3D Object Detection with Multiple Sensors}
There are several methods exploiting how to fuse information from multiple sensors for object detection. MV3D \cite{MV3D} projects LiDAR point cloud to bird-eye view (BEV) in order to generate proposals. These proposals with other information from image, front view and BEV are then sent to the second stage to predict final bounding boxes. AVOD \cite{AVOD} extends MV3D by introducing image features in the proposal generation stage. MMF \cite{Liang2019CVPR} fuses information from depth maps, LiDAR point clouds, images and maps to accomplish multiple tasks including depth completion, 2D object detection and 3D object detection. These tasks benefit each other and enhance final performance on 3D object detection.

\paragraph{3D Object Detection with LiDAR Only}
There are mainly two streams of methods dealing with 3D object detection only using LiDAR data. One is voxel-based, which applies voxelization on the entire point cloud. The difference among voxel-based methods lies on the initialization of voxel features. In \cite{VotetoVote}, each non-empty voxel is encoded with 6 statistical quantities by the points within this voxel. Binary encoding is used in \cite{FullyConvolutionNetworkForVehicle} for each voxel grid. VoxelNet \cite{VOXELNET} utilizes PointNet \cite{POINTNET} to extract features of each voxel. Compared to \cite{VOXELNET}, SECOND \cite{yan2018second} applies sparse convolution layers \cite{sparseconv} for parsing the compact representation. PointPillars \cite{lang2018pointpillars} treats pseudo-images as the representation after voxelization.

Another one is point-based. They take raw point cloud as input, and generate predictions based on each point. F-PointNet \cite{FPOINTNET} and IPOD \cite{yang2018ipod} adopt 2D mechanisms like detection or segmentation to filter most useless points, and generate predictions from these kept useful points. PointRCNN \cite{shi2018pointrcnn} utilizes a PointNet++ \cite{POINTNET2} with SA and FP layers to extract features for each point, proposes a region proposal network (RPN) to generate proposals, and applies a refinement module to predict bounding boxes and class labels. These methods outperform voxel-based, but their unbearable inference time makes it impossible to be applied in real-time autonomous driving system.
STD \cite{yang2019std} tries to take advantages of both point-based and voxel-based methods. It uses raw point cloud as input, applies PointNet++ to extract features, proposes a PointsPool layer for converting features from sparse to dense representations and utilizes CNNs in the refinement module. Although it is faster than all former point-based methods, it is still much slower than voxel-based methods. As analyzed before, all point-based methods are composed of two stages, which are proposal generation module including SA layers and FP layers, and a refinement module as the second stage for accurate predictions. In this paper, we propose to remove FP layers and the refinement module so as to speed up point-based methods.

\section{Our Framework}

%%Our framework is a single stage anchor-free 3D object detector, which has comparable accuracy to 2-stage point-based methods and is as fast as voxel-based methods. In order to break speed bottleneck, we remove FP layers and the second stage, which are used in all point-based detection methods, by proposing a novel sub-sampling strategy called fusion sampling so as to keep more positive points after SA layers. 
%%In the regression head, a modified SA layer and a center-ness assignment strategy are developed to speed up our model further and ensure its accuracy.

\begin{figure*}[ht]
  \centering
  \includegraphics[width=0.7\linewidth]{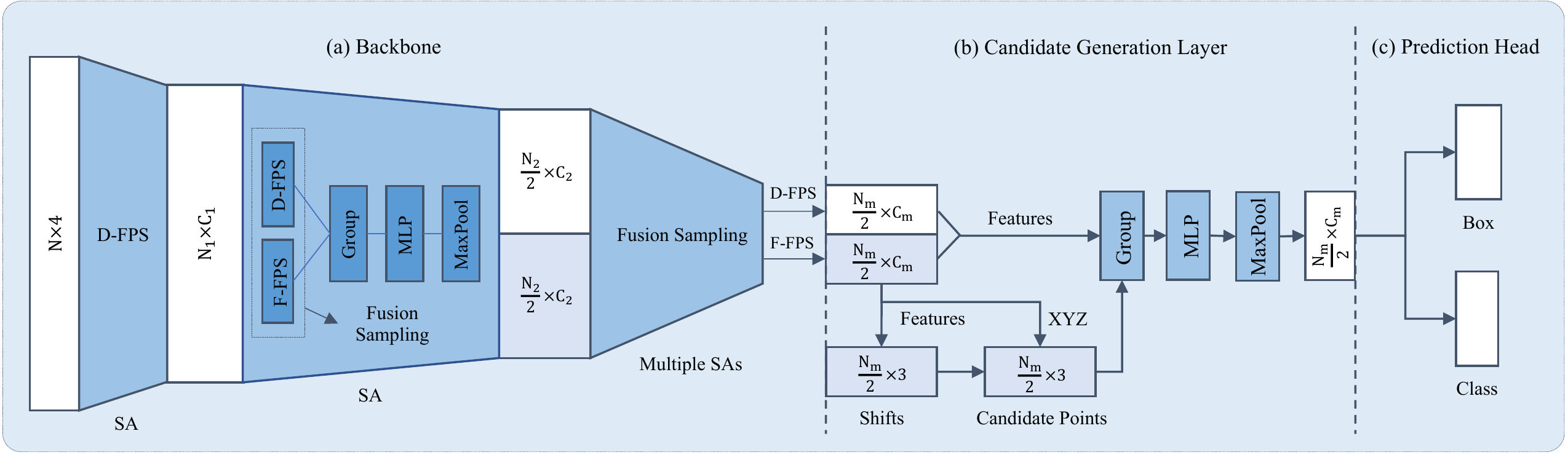}\\
  \caption{Illustration of the 3DSSD framework. On the whole, it is composed of backbone and box prediction network including a candidate generation layer and an anchor-free prediction head. (a) Backbone network. It takes the raw point cloud $(x, y, z, r)$ as input, and generates global features for all representative points through several SA layers with fusion sampling (FS) strategy. (b) Candidate generation layer (CG). It downsamples, shifts and extracts features for representative points after SA layers. (c) Anchor-free prediction head.}
  \label{fig:framework}
\end{figure*}

In this section, we first analyze the bottleneck of point-based methods, and describe our proposed fusion sampling strategy. Next, we present the box prediction network including a candidate generation layer, anchor-free regression head and our 3D center-ness assignment strategy. Finally, we discuss the loss function. The whole framework of 3DSSD is illustrated in Figure \ref{fig:framework}.

\subsection{Fusion Sampling}

\paragraph{Motivation} Currently, there are two streams of methods in 3D object detection, which are point-based and voxel-based frameworks. Albeit accurate, point-based methods are more time-consuming compared to voxel-based ones. We observe that all current point-based methods \cite{yang2019std,shi2018pointrcnn,yang2018ipod} are composed of two stages including proposal generation stage and prediction refinement stage. In first stage, SA layers are applied to downsample points for better efficiency and enlarging receptive fields while FP layers are applied to broadcast features for dropped points during downsampling process so as to recover all points. In the second stage, a refinement module optimizes proposals from RPN to get more accurate predictions. SA layers are necessary for extracting features of points, but FP layers and the refinement module indeed limit the efficiency of point-based methods, as illustrated in Table \ref{tab:part3_timetable}. Therefore, we are motivated to design a lightweight and effective point-based single stage detector.

\paragraph{Challenge} However, it is non-trivial to remove FP layers. %Because in point-based methods, predictions are derived from points, we have to make sure that at least one inner point for each instance in a scene is reserved after SA layers so as to guarantee all instances can be detected. 
As mentioned before, SA layers in backbone utilize D-FPS to choose a subset of points as the downsampled {\it representative points}. Without FP layers, the box prediction network is conducted on those surviving representative points.  Nonetheless, this sampling method only takes the relative locations among points into consideration. Consequently a large portion of surviving representative points are actually background points, like ground points, due to its large amount. In other words, there are several foreground instances which are totally erased through this process, making them impossible to be detected. %many points with similar semantic information are retained, like ground points, most of which are useless. 

With a limit of total representative points number $N_m$, for some remote instances, their inner points are not likely to be selected, because the amount of them is much smaller than that of background points. The situation becomes even worse on more complex datasets like nuScenes \cite{nuscenes2019} dataset. Statistically, we use points recall -- the quotient between the number of instances whose interior points survived in the sampled representative points and the total number of instances, to help illustrate this fact. As illustrated in the first row of Table \ref{tab:pr_on_nuscenes}, with 1024 or 512 representative points, their points recalls are only 65.9\% or 51.8\% respectively, which means nearly half of instances are totally erased, that is, cannot be detected. To avoid this circumstance, most of existing methods apply FP layers to recall those abandoned useful points during downsampling, but they have to pay the overhead of computation with longer inference time. 

\begin{table}[t]
   \centering \addtolength{\tabcolsep}{-1pt}
   \footnotesize
   \begin{tabular}{|c|c|c|c|}
       \hline
       Methods & SA layers (ms) & FP layers (ms) & Refinement Module (ms) \\
       \hline
       Baseline & 40 & 14 & 35 \\
      \hline
   \end{tabular}\vspace{0.1cm}
   \caption{Running time of difference components in our reproduced PointRCNN \cite{shi2018pointrcnn} model, which is composed of 4 SA layers and 4 FP layers for feature extraction, and a refinement module with 3 SA layers for prediction.}
   \label{tab:part3_timetable}
\end{table}  

\begin{table}[t]
   \centering \addtolength{\tabcolsep}{-1pt}
   \footnotesize
   \begin{tabular}{|c|c|c|c|}
       \hline
       Methods & 4096 & 1024 & 512 \\
       \hline
       D-FPS & 99.7 \% & 65.9 \% & 51.8 \% \\
       \hline
       F-FPS ($\lambda$=0.0) & 99.7 \% & 83.5 \% & 68.4 \% \\
       \hline
       F-FPS ($\lambda$=0.5) & 99.7 \% & 84.9 \% & 74.9 \% \\
       \hline
       F-FPS ($\lambda$=1.0) & 99.7 \% & 89.2 \% & 76.1 \% \\
       \hline
       F-FPS ($\lambda$=2.0) & 99.7 \% & 86.3 \% & 73.7 \% \\
       \hline
   \end{tabular}\vspace{0.1cm}
   \caption{Points recall among different sampling strategies on nuScenes dataset. ``4096'', ``1024'' and ``512'' represent the amount of representative points in the subset.}
   \label{tab:pr_on_nuscenes}
\end{table} 

\paragraph{Feature-FPS} 
%\tomua{For simplicity, in the later sections, we call interior points within any instance {\it positive points}, and the rest {\it negative points}.}
In order to preserve {\it positive points} (interior points within any instance) and erase those meaningless {\it negative points} (points locating on background), we have to consider not only spatial distance but also semantic information of each point during the sampling process. %which is represented by features in a deep neural network. 
We note that semantic information is well captured by the deep neural network. %Specifically, we treat the feature distance as part of the criterion in FPS, since they can tell the difference among points in a semantic level. 
So, utilizing the feature distance as the criterion in FPS, many similar useless negative points will be mostly removed, like massive of ground points. Even for positive points of remote objects, they can also get survived, because semantic features of points from different objects are distinct from each other. 

However, only taking the semantic feature distance as the sole criterion will preserve many points within a same instance, which introduces redundancy as well. For example, given a car, there is much difference between features of points around the windows and the wheels. As a result, points around the two parts will be sampled while any point in either part is informative for regression. To reduce the redundancy and increase the diversity, we apply both spatial distance and semantic feature distance as the criterion in FPS. It is formulated as
\begin{equation} \label{eq:ffps}
\begin{aligned}
C(A, B) = \lambda  L_d(A, B) + L_f(A, B),
\end{aligned}
\end{equation}
where $L_d(A, B)$ and $L_f(A, B)$ represent L2 XYZ distance and L2 feature distance between two points and $\lambda$ is the balance factor. We call this sampling method as Feature-FPS (F-FPS). The comparison among different $\lambda$ is shown in in Table \ref{tab:pr_on_nuscenes}, which demonstrates that combining two distances together as the criterion in the downsampling operation is more powerful than only using feature distance, \ie, the special case where $\lambda$ equals to 0. Moreover, as illustrated in Table \ref{tab:pr_on_nuscenes}, using F-FPS with 1024 representative points and $\lambda$  setting to 1 guarantees 89.2\% instances can be preserved in nuScenes \cite{nuscenes2019} dataset, which is 23.3\% higher than D-FPS sampling strategy.

\begin{figure}[bpt]
  \centering
  \includegraphics[width=1.0\linewidth]{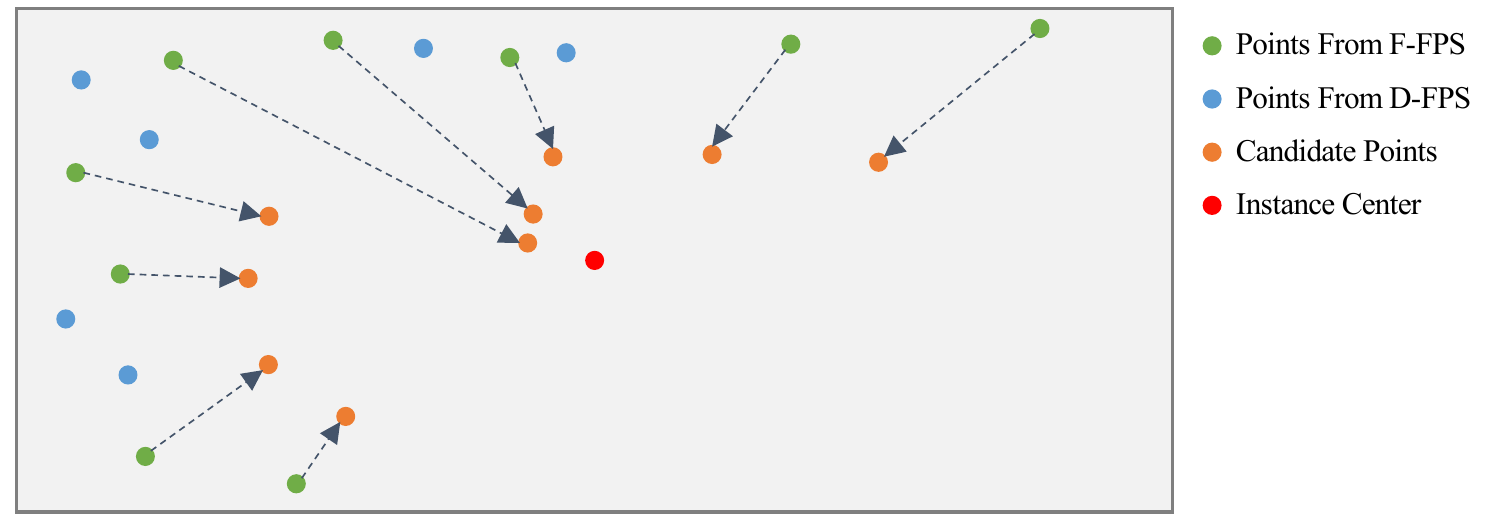}\\
  \caption{Illustration of the shifting operation in CG layer. The gray rectangle represents an instance with all positive representative points from F-FPS (green) and D-FPS (blue). The red dot represents instance center. We only shift points from F-FPS under the supervision of their distances to the center of an instance.}
  \label{fig:msa}
\end{figure}

\paragraph{Fusion Sampling} 
Large amount of positive points within different instances are preserved after SA layers thanks to F-FPS. However, with the limit of a fixed number of total representative points $N_m$, many negative points are discarded during the downsampling process, which benefits regression but hampers classification. That is, during the grouping stage in a SA layer, which aggregates features from neighboring points, a negative point is unable to find enough surrounding points, making it impossible to enlarge its receptive field. As a result, the model finds difficulty in distinguishing positive and negative points, leading to a poor performance in classification. Our experiments also demonstrate this limitation in ablation study. Although the model with F-FPS has higher recall rate and better localization accuracy than the one with D-FPS, it prefers treating many negative points as positive ones, leading to a drop in classification accuracy.

As analyzed above, after a SA layer, not only positive points should be sampled as many as possible, but we also need to gather enough negative points for more reliable classification. We present a novel fusion sampling strategy (FS), in which both F-FPS and D-FPS are applied during a SA layer, to retain more positive points for localization and keep enough negative points for classification as well. Specifically, we sample $\frac{N_m}{2}$ points respectively with F-FPS and D-FPS and feed the two sets together to the following grouping operation in a SA layer.

\begin{figure*}[bpt]
  \centering
  \includegraphics[width=1.0\linewidth]{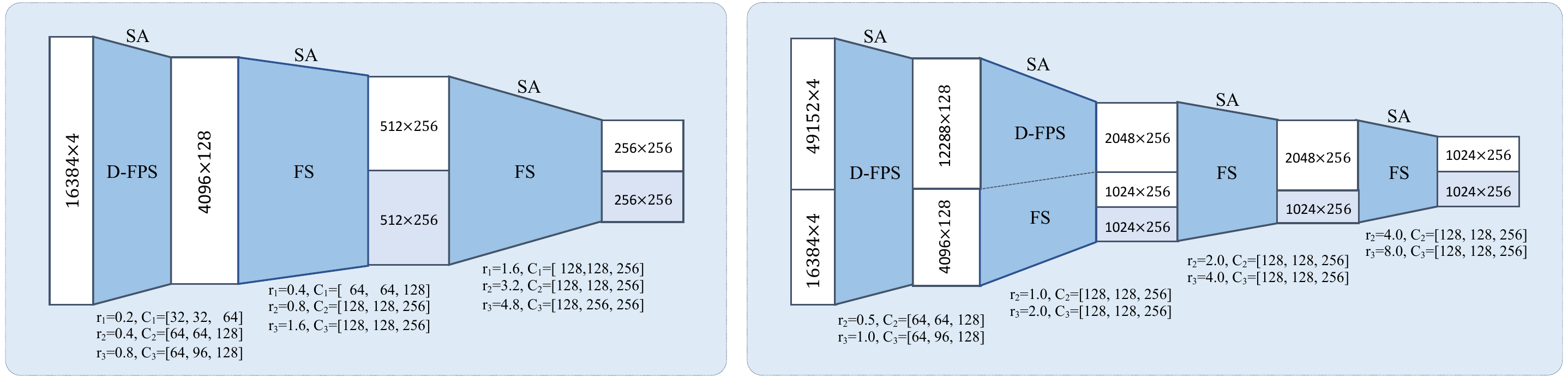}\\
  \caption{Backbone network of 3DSSD on KITTI (left) and nuScenes (right) datasets.}
  \label{fig:backbone}
\end{figure*}

\subsection{Box Prediction Network}

\paragraph{Candidate Generation Layer}
After the backbone network implemented by several SA layers intertwined with fusion sampling, we gain a subset of points from both F-FPS and D-FPS, which are used for final predictions. In former point-based methods, another SA layer should be applied to extract features before the prediction head. There are three steps in a normal SA layer, including center point selection, surrounding points extraction and semantic feature generation.

In order to further reduce computation cost and fully utilize the advantages of fusion sampling, we present a candidate generation layer (CG) before our prediction head, which is a variant of SA layer. Since most of representative points from D-FPS are negative points and useless in bounding box regression, we only use those from F-FPS as initial center points. These initial center points are shifted under the supervision of their relative locations to their corresponding instances as illustrated in Figure \ref{fig:msa}, same as VoteNet \cite{Qi2019Votenet}. We call these new points after shifting operation as {\it candidate points}. Then we treat these candidate points as the center points in our CG layer. We use candidate points rather than original points as the center points for the sake of performance, which will be discussed in detail later. Next, we find the surrounding points of each candidate point from the whole representative point set containing points from both D-FPS and F-FPS with a pre-defined range threshold, concatenate their normalized locations and semantic features as input, and apply MLP layers to extract features. These features will be sent to the prediction head for regression and classification. This entire process is illustrated in Figure \ref{fig:framework}.

\paragraph{Anchor-free Regression Head} 
With fusion sampling strategy and the CG layer, our model can safely remove the time-consuming FP layers and the refinement module. In the regression head, we are faced with two options, anchor-based or anchor-free prediction network. If anchor-based head is adopted, we have to construct multi-scale and multi-orientation anchors so as to cover objects with variant sizes and orientations. Especially in complex scenes like those in the nuScenes dataset \cite{nuscenes2019}, where objects are from 10 different categories with a wide range of orientations, we need at least 20 anchors, including 10 different sizes and 2 different orientations $(0, \pi / 2)$ in an anchor-based model. To avoid the cumbersome setting of multiple anchors and be consistent with our lightweight design, we utilize anchor-free regression head.
%Moreover, given that point cloud is part-visible, different anchors may have same features because they have same interior points, which will lead to ambiguous during training for they have different regression labels.

In the regression head, for each candidate point, we predict the distance  $(d_x, d_y, d_z)$ to its corresponding instance, as well as the size $(d_l, d_w, d_h)$ and orientation of its corresponding instance.
%The first reason of using anchor-free head is that, it could save a lot of computaions during inference, which boosts our model even faster. The second reason is that, point cloud can provide more structual information, which means it is aware of distance. Therefore, it is a better choice to directly predict distance between points and groundtruth, rather than predict these encoding values from distance and anchor sizes. 
Since there is no prior orientation of each point, we apply a hybrid of classification and regression formulation following \cite{FPOINTNET} in orientation angle regression. Specifically, we pre-define $N_a$ equally split orientation angle bins and classify the proposal orientation angle into different bins. Residual is regressed with respect to the bin value. $N_a$ is set to 12 in our experiments.

\paragraph{3D Center-ness Assignment Strategy}\label{ctr_aware_layer}
In the training process, we need an assignment strategy to assign labels for each candidate point. In 2D single stage detectors, they usually use intersection-over-union (IoU) \cite{SSD} threshold or mask \cite{FCOS,Yangze2019Reppoint} to assign labels for pixels. FCOS \cite{FCOS} proposes a continuous center-ness label, replacing original binary classification label, to further distinguish pixels. It assigns higher center-ness scores to pixels closer to instance centers, leading to a relatively better performance compared to IoU- or mask-based assignment strategy. However, it is unsatisfying to directly apply center-ness label to 3D detection task. Given that all LiDAR points are located on surfaces of objects, their center-ness labels are all very small and similar, which makes it impossible to distinguish good predictions from other points.

%For points sent to this m, it will generate geometrical shifts $(\delta_x, \delta_y, \delta_z)$ by their semantic features and translate them by adding these shifts. These shifts are supervised by centers of their corresponding groundtruth during training. These translated points are then sent to the anchor-free regression head for final predictions.

%Before constructing 3D center-ness label, we predict shifts $(\delta_x, \delta_y, \delta_z)$ for the sampled points to generate candidate points, supervised by the distances between their locations and the centers of their corresponding instances in our CG layer as stated above. 

Instead of utilizing original representative points in point cloud, we resort to the predicted candidate points, which are supervised to be close to instance centers. Candidate points closer to instance centers tend to get more accurate localization predictions, and 3D center-ness labels are able to distinguish them easily. For each candidate point, we define its center-ness label through two steps. We first determine whether it is within an instance $l_{mask}$, which is a binary value. Then we draw a center-ness label according to its distance to 6 surfaces of its corresponding instance. The center-ness label is calculated as 
\begin{equation} \label{eq:center_ness}
\begin{aligned}
l_{ctrness} = \sqrt[3]{\frac{min(f, b)}{max(f, b)} \times \frac{min(l, r)}{max(l, r)} \times \frac{min(t, d)}{max(t, d)}},
\end{aligned}
\end{equation}
where $(f, b, l, r, t, d)$ represent the distance to front, back, left, right, top and bottom surfaces respectively. The final classification label is the multiplication of $l_{mask}$ and $l_{ctrness}$.

\subsection{Loss Function}

The overall loss is composed of classification loss, regression loss and shifting loss, as 
\begin{equation} \label{eq:total_loss}
\begin{aligned}
L = &\frac{1}{N_{c}} \sum_i L_{c} (s_i, u_i) + \lambda_1 \frac{1}{N_{p}} \sum_i  [u_i > 0] L_{r} \\
&+ \lambda_2 \frac{1}{N_{p}^{*}} L_{s},
\end{aligned}
\end{equation}
where $N_{c}$ and $N_{p}$ are the number of total candidate points and positive candidate points, which are candidate points locating in foreground instance. In the classification loss, we denote $s_i$ and $u_i$ as the predicted classification score and center-ness label for point $i$ respectively and use cross entropy loss as $L_{c}$. 

The regression loss $L_{r}$ includes distance regression loss $L_{dist}$, size regression loss $L_{size}$, angle regression loss $L_{angle}$ and corner loss $L_{corner}$. Specifically, we utilize the smooth-$l_1$ loss for $L_{dist}$ and $L_{size}$, in which the targets are offsets from candidate points to their corresponding instance centers and sizes of corresponding instances respectively. Angle regression loss contains orientation classification loss and residual prediction loss as
\begin{equation} \label{eq:anglereg}
L_{angle} = L_{c}(d_{c}^{a}, t_{c}^a) + D(d_{r}^a, t_{r}^a),
\end{equation}
where $d_{c}^a$ and $d_{r}^a$ are predicted angle class and residual while $t_{c}^a$ and $t_{r}^a$ are their targets. Corner loss is the distance between the predicted 8 corners and assigned ground-truth, expressed as 
\begin{equation} \label{eq:cornerloss}
L_{corner} = \sum_{m=1}^8 \norm{P_{m} - G_{m}},
\end{equation}
where $P_{m}$ and $G_{m}$ are the location of ground-truth and prediction for point $m$.  

As for the shifting loss $L_{s}$, which is the supervision of shifts prediction in CG layer, we utilize a smooth-$l_1$ loss to calculate the distance between the predicted shifts and the residuals from representative points to their corresponding instance centers. $N_{p}^*$ is the amount of positive representative points from F-FPS.

\section{Experiments}

We evaluate our model on two datasets: the widely used KITTI Object Detection Benchmark \cite{KITTIDATASET1,KITTIDATASET2}, and a larger and more complex nuScenes dataset \cite{nuscenes2019}. 

\subsection{KITTI}
There are 7,481 training images/point clouds and 7,518 test images/point clouds with three categories of Car, Pedestrian and Cyclist in the KITTI dataset. We only evaluate our model on class Car, due to its large amount of data and complex scenarios. Moreover, most of state-of-the-art methods only test their models on this class. We use average precision (AP) metric to compare with different methods. During evaluation, we follow the official KITTI evaluation protocol -- that is, the IoU threshold is 0.7 for class Car.

\begin{table}[t]
   \centering 
   \footnotesize
   \begin{tabular}{|c|c|c||c|c|c|}
       \hline
       \multicolumn{1}{|c|}{ \multirow{2}{*}{Type}} & \multicolumn{1}{c|}{ \multirow{2}{*}{Method}} & \multicolumn{1}{c||}{ \multirow{2}{*}{Sens.}} & \multicolumn{3}{|c|}{$AP_{3D} (\%)$} \\ \cline{4-6}
       \multicolumn{1}{|c|}{} & \multicolumn{1}{c|}{} & \multicolumn{1}{c||}{} & \multicolumn{1}{|c|}{Easy} & \multicolumn{1}{|c|}{Mod} & \multicolumn{1}{|c|}{Hard} \\
       \hline
       \hline
      \multirow {12}{*}{2-stage} & MV3D \cite{MV3D} & \multirow{7}{*}{R + L} & 71.09 & 62.35 & 55.12\\
      {} & AVOD \cite{AVOD} & {} & 76.39 & 66.47  & 60.23 \\
      {} & F-PointNet \cite{FPOINTNET} & {} & 82.19  & 69.79  & 60.59 \\ 
      {} & AVOD-FPN \cite{AVOD} & {} & 83.07  & 71.76  & 65.73 \\
      {} & IPOD \cite{yang2018ipod} & {} & 80.30  & 73.04  & 68.73 \\
      {} & UberATG-MMF \cite{Liang2019CVPR} & {} & 88.40  & 77.43  & 70.22 \\ 
      {} & F-ConvNet \cite{wang2019frustum} & {} & 87.36 & 76.39 & 66.69 \\ \cline{2-6}
      {} & PointRCNN \cite{shi2018pointrcnn} & \multirow{5}{*}{L} & 86.96  & 75.64  & 70.70 \\
      {} & Fast Point-RCNN \cite{Chen2019fastpointrcnn} & {} & 85.29  & 77.40  & 70.24 \\
      {} & Patches \cite{lehner2019patch} & {} & {\color{red} 88.67}  & 77.20  & 71.82 \\
      {} & MMLab-PartA\^{}2 \cite{shi2019part} & {} & 87.81 & 78.49 & 73.51 \\
      {} & STD \cite{yang2019std} & {} & 87.95  & {\color{red} 79.71}  & {\color{red} 75.09} \\
      \hline
      \hline
      \multirow {5}{*}{1-stage} & ContFuse \cite{CONTFUSE} & \multirow{1}{*}{R + L} & 83.68  & 68.78  & 61.67 \\ \cline{2-6}
      {} & VoxelNet \cite{VOXELNET} & \multirow{4}{*}{L} & 77.82  & 64.17  & 57.51 \\ 
      {} & SECOND \cite{yan2018second} & {} & 84.65  & 75.96  & 68.71 \\
      {} & PointPillars \cite{lang2018pointpillars} & {} & 82.58  & 74.31  & 68.99 \\
      {} & Ours & {} & \bf 88.36  & \bf 79.57  & \bf 74.55 \\
      \hline
   \end{tabular}\vspace{0.1cm}
   \caption{Results on KITTI test set on class Car drawn from official Benchmark \cite{KITTI3DBENCHMARK}. ``Sens.'' means sensors used by the method. ``L'' and ``R'' represent using LiDAR and RGB images respectively.}\label{tab:mainkitti}
\end{table}

\vspace{-0.05in}
\paragraph{Implementation Details}

To align network input, we randomly choose 16k points from the entire point cloud per scene. The detail of backbone nework is illustrated in Figure \ref{fig:backbone}.  The network is trained by ADAM \cite{AdamOptimizer} optimizer with an initial learning rate of 0.002 and a batch size of 16 equally distributed on 4 GPU cards. The learning rate is decayed by 10 at 40 epochs. We train our model for 50 epochs.

We adopt 4 different data augmentation strategies on KITTI dataset in order to prevent overfitting. First, we use mix-up strategy same as \cite{yan2018second}, which randomly adds foreground instances with their inner points from other scenes to current point cloud. Then, for each bounding box, we rotate it following a uniform distribution $\Delta \theta_1 \in [- \pi / 4, + \pi / 4]$ and add a random translation ($\Delta x, \Delta y, \Delta z$). Third, each point cloud is randomly flipped along $x$-axis. Finally, we randomly rotate each point cloud around $z$-axis (up axis) and rescale it.

\begin{table*}[t]
   \centering 
   \footnotesize
   \begin{tabular}{|c|c|c|c|c|c|c|c|c|c|c|c|}
      \hline
      & Car & Ped & Bus & Barrier & TC & Truck & Trailer & Moto & Cons. Veh. & Bicycle & mAP \\
      \hline
      SECOND \cite{yan2018second} & 75.53 & 59.86 & 29.04 & 32.21 & 22.49 & 21.88 & 12.96 & 16.89 & 0.36 & 0 & 27.12 \\
      \hline
      PointPillars \cite{lang2018pointpillars} & 70.5 & 59.9 & 34.4 & 33.2  & 29.6 & 25.0 & 20.0 & 16.7 & 4.5  & 1.6 & 29.5 \\
      \hline 
      Ours & \bf 81.20 & \bf 70.17 & \bf 61.41 & \bf 47.94 & \bf 31.06 & \bf 47.15 & \bf 30.45 & \bf 35.96 & \bf 12.64 & \bf 8.63 & \bf 42.66 \\
      \hline
   \end{tabular}\vspace{0.1cm}
   \caption{AP on nuScenes dataset. The results of SECOND come from its official implementation \cite{SecondCode}. }\label{tab:ap_nuscenes}
   %}
\end{table*}

\begin{table}[t]
   \centering \addtolength{\tabcolsep}{-1pt}
   \footnotesize
   \begin{tabular}{|c|c|c|c|c|c|c|c|}
      \hline
      & mAP & mATE & mASE & mAOE & mAVE & AAE & NDS \\
      \hline
      PP \cite{lang2018pointpillars} & 29.5 & 0.54 & \bf 0.29 & 0.45 & 0.29 & 0.41 & 44.9 \\
      \hline 
      Ours & \bf 42.6 & \bf 0.39 & \bf 0.29 & \bf 0.44 & \bf 0.22 & \bf 0.12 & \bf 56.4 \\
      \hline
   \end{tabular}\vspace{0.1cm}
   \caption{NDS on nuScenes dataset. ``PP'' represents PointPillars.}\label{tab:nds_nuscenes}
\end{table}

\vspace{-0.05in}
\paragraph{Main Results}

\begin{figure*}[t]
  \centering
  \includegraphics[width=0.88\linewidth]{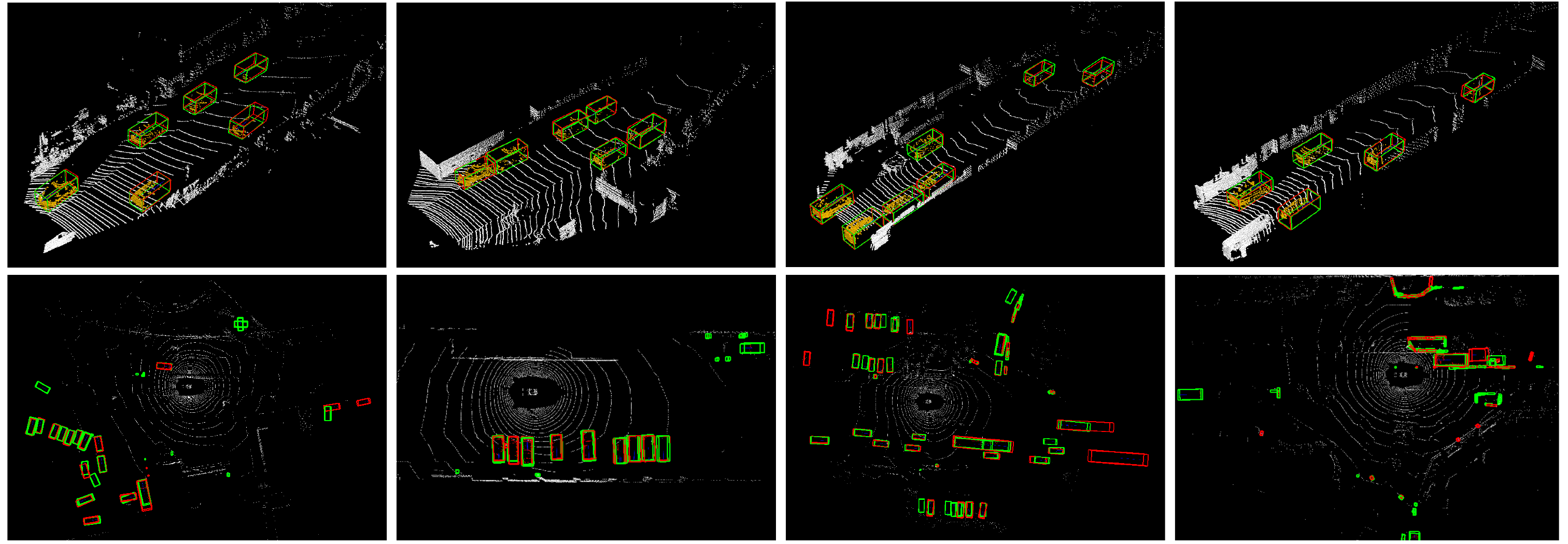}\\
  \caption{Qualitative results of 3DSSD on KITTI (top) and nuScenes (bottom) dataset. The groundtruths and predictions are labeled as red and green respectively.}
  \label{fig:results}
\end{figure*}

As illustrated in Table \ref{tab:mainkitti}, our method outperforms all state-of-the-art voxel-based single stage detectors by a large margin. On the main metric, \ie, AP on ``moderate'' instances, our method outperforms SECOND \cite{yan2018second} and PointPillars \cite{lang2018pointpillars} by $3.61 \%$ and $5.26 \%$ respectively. Still, it retains comparable performance to the state-of-the-art point-based method STD \cite{yang2019std} with a more than 2$\times$ faster inference time. Our method still outperforms other two stage methods like part-A\^{}2 net and PointRCNN by $1.08 \%$ and $3.93 \%$ respectively. Moreover, it also shows its superiority when compared to multi-sensors methods, like MMF \cite{Liang2019CVPR} and F-ConvNet \cite{wang2019frustum}, which achieves about $2.14 \%$ and $3.18 \%$ improvements respectively. We present several qualitative results in Figure \ref{fig:results}.

\subsection{nuScenes}

The nuScenes dataset is a more challenging dataset.
It contains 1000 scenes, gathered from Boston and Singapore due to their dense traffic and highly challenging driving situations. It provides us with 1.4M 3D objects on 10 different classes, as well as their attributes and velocities. There are about 40k points per frame, and in order to predict velocity and attribute, all former methods combine points from key frame and frames in last 0.5s, leading to about 400k points. Faced with such a large amount of points, all point-based two stage methods perform worse than voxel-based methods on this dataset, due to the GPU memory limitation. In the benchmark, a new evaluation metric called nuScenes detection score (NDS) is presented, which is a weighted sum between mean average precision (mAP), the mean average errors of location (mATE), size (mASE), orientation (mAOE), attribute (mAAE) and velocity (mAVE). We use $TP$ to denote the set of the five mean average errors, and NDS is calculated by 

\begin{equation} \label{eq:nds}
NDS = \frac{1}{10} [5 mAP + \sum_{mTP \in TP}(1 - min(1, mTP)) ].
\end{equation}

\vspace{-0.05in}
\paragraph{Implementation Details} 

For each key frame, we combine its points with points in frames within last 0.5s so as to get a richer point cloud input, just the same as other methods. Then, we apply voxelization for randomly sampling point clouds, so as to align input as well as keep original distribution. We randomly choose 65536 voxels, including 16384 voxels from key frame and 49152 voxels from other frames. The voxel size is $[0.1, 0.1, 0.1]$, and 1 interior point is randomly selected from each voxel. We feed these 65536 points into our point-based network. 

The backbone network is illustrated in Figure \ref{fig:backbone}. The training schedule is just the same as the schedule on KITTI dataset. We only apply flip augmentation during training.

\vspace{-0.05in}
\paragraph{Main results}
We show NDS and mAP among different methods in Table \ref{tab:nds_nuscenes}, and compare their APs of each class in Table \ref{tab:ap_nuscenes}.
As illustrated in Table \ref{tab:nds_nuscenes}, our method draws better performance compared to all voxel-based single stage methods by a large margin. Not only on mAP, it also outperforms those methods on  AP of each class, as illustrated in Table \ref{tab:ap_nuscenes}. 
The results show that our model can deal with different objects with a large variance on scale. Even for a huge scene with many negative points, our fusion sampling strategy still has the ability to gather enough positive points out. 
In addition, better results on velocity and attribute illustrate that our model can also discriminate information from different frames.

%Note that, on nuscenes test leaderboard, \cite{zhu2019cbalance} ranked the 1st place. However, it uses DB Sampling, which adds more class-balanced training samples to its training set, for enlarging the training set. As a result, we will not compare our method to them for using different training set. 

\subsection{Ablation Studies}

All ablation studies are conducted on KITTI dataset \cite{KITTIDATASET1}. We follow VoxelNet \cite{VOXELNET} to split original training set to 3,717 images/scenes train set and 3,769 images/scenes val set. All ``AP'' results in ablation studies are calculated on ``Moderate'' difficulty level.

\begin{table}[t]
   \centering \addtolength{\tabcolsep}{-1pt}
   \footnotesize
   \begin{tabular}{|c|c|c|c|}
      \hline
      Method & Easy & Moderate & Hard \\
      \hline
      VoxelNet \cite{VOXELNET} & 81.97 & 65.46 & 62.85 \\ 
      \hline
      SECOND \cite{yan2018second} & 87.43 & 76.48 & 69.10 \\
      \hline
      PointPillars \cite{lang2018pointpillars} & - & 77.98 & - \\
      \hline
      Ours & \bf 89.71 & \bf 79.45 & \bf 78.67 \\
      \hline
   \end{tabular}\vspace{0.1cm}
   \caption{3D detection AP on KITTI val set of our model for ``Car" compared to other state-of-the-art single stage methods.}\vspace{-0.05in}
   \label{tab:kittival_compare}
\end{table}

\begin{figure*}[t]
  \centering
  \includegraphics[width=0.88\linewidth]{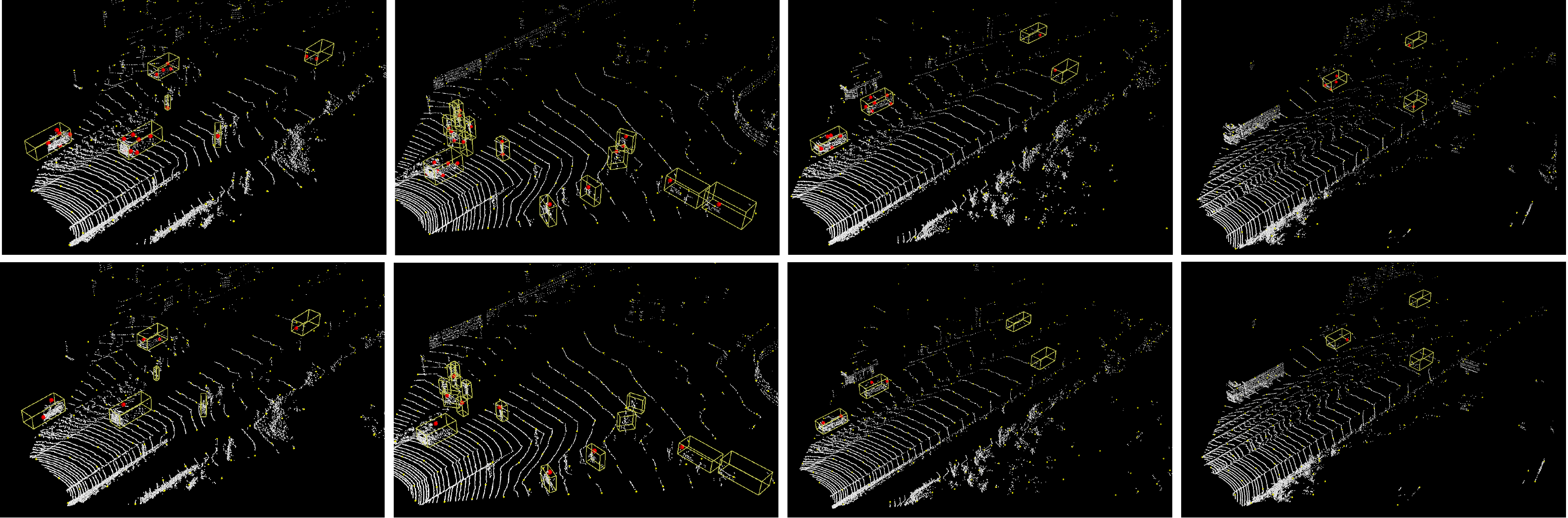}\\
  \caption{Comparision between representative points after fusion sampling (top) and D-FPS only (bottom). The whole point cloud and all representative points are colored in white and yellow respectively. Red points represent positive representative points.}
  \label{fig:ffps_compare}
\end{figure*}

\vspace{-0.05in}
\paragraph{Results on Validation Set}
We report and compare the performance of our method on KITTI validation set with other state-of-the-art voxel-based single stage methods in Table \ref{tab:kittival_compare}. As shown, on the most important ``moderate'' difficulty level, our method outperforms by $1.47 \%$, $2.97 \%$ and $13.99 \%$ compared to PointPillars, SECOND and VoxelNet respectively. This illustrates the superiority of our method.

\vspace{-0.05in}
\paragraph{Effect of Fusion Sampling Strategy} 
Our fusion sampling strategy is composed of F-FPS and D-FPS. We compare points recall and AP among different sub-sampling methods in Table \ref{tab:pts_recall}. Sampling strategies containing F-FPS have higher points recall than D-FPS only. In Figure \ref{fig:ffps_compare}, we also present some visual examples to illustrate the benefits of F-FPS to fusion sampling. In addition, the fusion sampling strategy receives a much higher AP, \ie, $2.7 \%$ better than the one with F-FPS only. The reason is that fusion sampling method can gather enough negative points, which enlarges receptive fields and achieve accurate classification results. 

%  Whether mask table
\begin{table}[t]
   \centering \addtolength{\tabcolsep}{-1pt}
   \footnotesize
   \begin{tabular}{|c|c|c|c|}
       \hline
        & D-FPS & F-FPS & FS \\
       \hline
       recall (\%) & 92.47 & \bf 98.45 & 98.31 \\
       \hline
       AP (\%) & 70.4 & 76.7 & \bf 79.4 \\
      \hline
   \end{tabular}\vspace{0.1cm}
   \caption{Points recall and AP from different sampling methods.}
   \label{tab:pts_recall}
\end{table}

%  Whether mask table
\begin{table}[t]
   \centering \addtolength{\tabcolsep}{-1pt}
   \footnotesize
   \begin{tabular}{|c|c|c|c|c|}
       \hline
       & IoU & Mask & 3D center-ness\\
       \hline
       without shifting (\%) & 70.4 & 76.1 & 43.0 \\
       \hline
       with shifting (\%) & 78.1 & 77.3 & \bf 79.4 \\
      \hline
   \end{tabular}\vspace{0.1cm}
   \caption{AP among different assignment strategies. ``with shifting'' means using shifts in the CG layer.}
   %\caption{AP between ``without shifting'' and ``with shifting'' from different assignment strategies. ``with shifting'' means using shifts in the CG layer.}
   \label{tab:assignment_strategy}
\end{table}

%  Whether mask table
\begin{comment}
\begin{table}[t]
   \centering \addtolength{\tabcolsep}{-1pt}
   \footnotesize
   \begin{tabular}{|c|c|c|c|}
       \hline
       distance & $d_3$ & $d_2$ & $d_1$ \\
       \hline
       AP (\%) & 76.7 & 77.4 & 78.7 \\
       \hline
       \hline
       center-ness & [0.0, 1.0] & [0.5, 1.0] & [1.0, 1.0] \\
       \hline
       AP (\%) & 78.8 & 79.4 & 76.7 \\
       \hline
   \end{tabular}\vspace{0.3cm}
   \caption{AP from distance strategy and center-ness strategies with different threshold.}
   \label{tab:assignment_strategy}
\end{table}
\end{comment}

\vspace{-0.05in}
\paragraph{Effect of Shifting in CG Layer}
In Table \ref{tab:assignment_strategy}, we compare performance between with or without shifting representative points from F-FPS in CG layer. Under different assignment strategies, APs of models with shifting are all higher than those without shifting operations. If the candidate points are closer to the centers of instances, it will be easier for them to retrieve their corresponding instances.

\vspace{-0.05in}
\paragraph{Effect of 3D Center-ness Assignment}
We compare the performance of different assignment strategies including IoU, mask and 3D center-ness label. As shown in Table \ref{tab:assignment_strategy}, with shifting operation, the model using center-ness label gains better performance than the other two strategies.

\begin{table}[t]
   \centering \addtolength{\tabcolsep}{-1pt}
   \footnotesize
   \begin{tabular}{|c|c|c|c|c|}
       \hline
       & F-PointNet \cite{FPOINTNET} & PointRCNN \cite{shi2018pointrcnn} & STD\cite{yang2019std} & Ours\\
       \hline
      time (ms) & 170 & 100 & 80 & \bf 38 \\
      \hline
   \end{tabular}\vspace{0.1cm}
   \caption{Inference time among different point-based methods.}
   \label{tab:inference_time}
\end{table}

\vspace{-0.05in}
\paragraph{Inference Time}
The total inference time of 3DSSD is 38ms, tested on KITTI dataset with a Titan V GPU. We compare inference time between 3DSSD and all existing point-based methods in Table \ref{tab:inference_time}. As illustrated, our method is much faster than all existing point-based methods. Moreover, our method maintains similar inference speed compared to state-of-the-art voxel-based single stage methods like SECOND which is 40ms. Among all existing methods, it is only slower than PointPillars, which is enhanced by several implementation optimizations such as TensorRT, which can also be utilized by our method for even faster inference speed.

\section{Conclusion}

In this paper, we first propose a lightweight and efficient point-based 3D single stage object detection framework. We introduce a novel fusion sampling strategy to remove the time-consuming FP layers and the refinement module in all existing point-based methods.
In the prediction network, a candidate generation layer is designed to reduce computation cost further and fully utilize downsampled representative points, and an anchor-free regression head with 3D center-ness label is proposed in order to boost the performance of our model. All of above delicate designs enable our model to be superiority to all existing single stage 3D detectors in both performance and inference time.

{\small
\bibliographystyle{ieee_fullname}
\bibliography{egbib}
}

\end{document}